\documentclass[10pt,a5paper,twoside]{article}
\usepackage{coling2012}

\newtheorem{algorithm}{Algorithm}

\usepackage{epstopdf}
\usepackage{subfigure}
\usepackage{epstopdf}
\usepackage{CJKutf8}
\usepackage{algorithm}
\usepackage{algorithmic}
\usepackage{multirow}
\usepackage{multirow}

\title{Domain-Specific Sentiment Word Extraction by Seed Expansion and Pattern Generation}
\author{$TANG~Du Yu^{1}~~~QIN~Bing^{1}~~~ZHOU~Lan Jun^{2}$\\$~~~WONG~Kam Fai^{2}~~~ZHAO~Yan Yan^{1}~~~LIU~Ting^{1}$\\
{\small  	(1) Research Center for Social Computing and Information Retrieval, Harbin Institute of Technology\\
 		(2) Dept. of Systems Engineering \& Engineering Management, The Chinese University of Hong Kong\\
  \texttt{\{dytang, qinb, zyy, tliu\}@ir.hit.edu.cn, \{ljzhou, kfwong\}@se.cuhk.edu.hk} \\
}}
\begin{document}
\maketitle




\abstractEn{
This paper focuses on the automatic extraction of domain-specific sentiment word (DSSW), which is a fundamental subtask of sentiment analysis. Most previous work utilizes manual patterns for this task. However, the performance of those methods highly relies on the labelled patterns or selected seeds. In order to overcome the above problem, this paper presents an automatic framework to detect large-scale domain-specific patterns for DSSW extraction. To this end, sentiment seeds are extracted from massive dataset of user comments. Subsequently, these sentiment seeds are expanded by synonyms using a bootstrapping mechanism. Simultaneously, a synonymy graph is built and the graph propagation algorithm is applied on the built synonymy graph. Afterwards, syntactic and sequential relations between target words and high-ranked sentiment words are extracted automatically to construct large-scale patterns, which are further used to extracte DSSWs. The experimental results in three domains reveal the effectiveness of our method.\\\\}

\hrule
\hrule

\newpage

\section{Introduction}

In recent years, sentiment analysis (or opinion mining) has attracted a lot of attention in natural language processing and information retrieval \cite{Pang2008}. An important subtask in sentiment analysis is building sentiment lexicons, which is fundamental for many sentiment analysis tasks, such as document-level \cite{Turney2002} and sentence-level \cite{Zhou2011} sentiment classification, collocation polarity disambiguation \cite{Zhao2012} and opinion retrieval \cite{Li2010}. Sentiment words, such as \emph{good}, \emph{bad}, \emph{excellent} and \emph{awful}, can indicate the sentiment polarity of text directly. However, sentiment words are domain-specific, because opinion expressions vary greatly in different domains \cite{Liu2012a}. A positive word in one domain may be neutral or negative in another domain. For example, "low" in "low cost" is positive but negative in "low salary". Therefore, it's necessary to extract domain-specific sentiment word (DSSWs) in different domains based on domain-specific text.

Recently, some methods are proposed for sentiment word extraction, including thesaurus-based \cite{Baccianella2010} and corpus-based \cite{Qiu2011} method. The performance of their algorithms more or less depends on the quality of labelled resources, selected seeds or manual patterns. Moreover, annotating each domain of interest is costly and time consuming. Thus, it's appealing to extract DSSWs utilizing the corpora without expensive labelling.

This paper presents an automatic framework for DSSW extraction. The advantage of our framework is to leverage domain-independent knowledge to detect large-scale syntactic and sequential patterns, which can be used to extract DSSWs. 
In the first step, high-quality sentiment seeds are selected from massive dataset of user comments. Then, sentiment seeds are expanded by synonyms in a bootstrapping schema, and a synonymy graph is built simultaneously. After that, graph propagation algorithm is applied on the synonymy graph to select general sentiment words. Finally, the syntactic and sequential relations between general sentiment words and target words are utilized to extract hundreds of patterns, which are used to extract DSSWs. Our approach differs from existing approaches in that it requires no labelled information except for the massive dataset of user comments. Thus, our proposed method can be viewed as an semi-supervised method. We test our method in three different domains, and our approach can extract accurate DSSWs from the target domain without annotated data. We also compare our method with two solid baseline methods, the experimental results demonstrate that our method outperforms them substantially.

Specifically, the contributions of this paper are as follows:
\begin{itemize}
\item This paper presents an automatic method to generate hundreds of domain-specific patterns for sentiment word extraction.
\item A simple and effective framework is proposed to extract DSSWs without any labelling.
\item This paper presents the first work on combining syntactic and sequential patterns for sentiment lexicon extraction.
\item The experimental results illustrate that our proposed method works effectively and outperforms two baselines largely.
\end{itemize}

\section{Method Overview}\label{motivation}
\begin{figure}[h]
\centering
\subfigure[example 1]{\label{example-1} \includegraphics[scale=0.7]{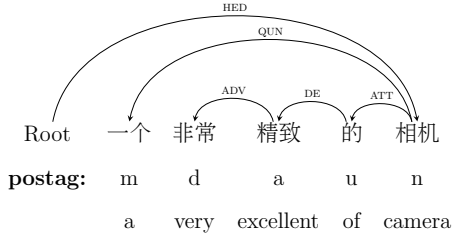} }

\subfigure[example 2]{\label{example-2} \includegraphics[scale=0.7]{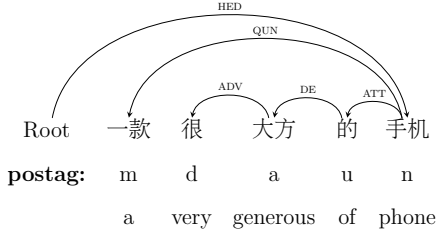} }

\vspace{-0.5em}
\caption{Example of dependency structure in digital domain.}

\label{figure-motivation}
\end{figure}

\begin{CJK*}{UTF8}{gbsn}
This section presents the brief idea behind our framework. Figure \ref{figure-motivation} shows two examples in digital domain after POS tagging and dependency parsing. In Figure \ref{example-1}, \emph{excellent[精致]} is a sentiment seed due to its stable polarity in different domains. Our goal is to extract new sentiment word (\emph{generous[大方]}) as DSSW by its structured similarity with general sentiment word (\emph{excellent}) when they are used to modify target words (such as \textit{phone} and \textit{camera}).
\end{CJK*}

In our framework, the first step is to select sentiment seeds, such as \emph{excellent}. The assumption is that sentiment seeds are used to modify diverse products with consistent polarity. Thus, we utilize massive dataset of user comments to extract high-confidence sentiment seeds. Afterwards, in order to get more general sentiment words, a bootstrapping mechanism is used to expand the sentiment seeds. At the same time, a synonymy graph is built and propagation algorithm is then utilized to calculate the sentiment of each word in the synonymy graph. As a result, high ranked words are selected as general sentiment words.

Then, general sentiment words are leveraged to extract DSSWs. In Figure \ref{example-1} and \ref{example-2}, it's obvious that there are some shared structures between sentiment words (e.g. \emph{excellent}, \emph{generous}) and target words (e.g. \emph{phone}, \emph{camera}). Thus, general sentiment words and these common structures can be used to extract new sentiment words. This paper extracts large-scale patterns to describe these structures, namely syntactic and sequential patterns. 
The detail of the method is described in Section \ref{methodology}.
\section{Methodology}\label{methodology}
\subsection{Framework Architecture}\label{framework overview}
The architecture of out framework is illustrated in Figure \ref{figure-framework}.
According to the framework, three components are carried out to extract DSSW:
\begin{figure}[h]
\centering
\includegraphics[width=\columnwidth]{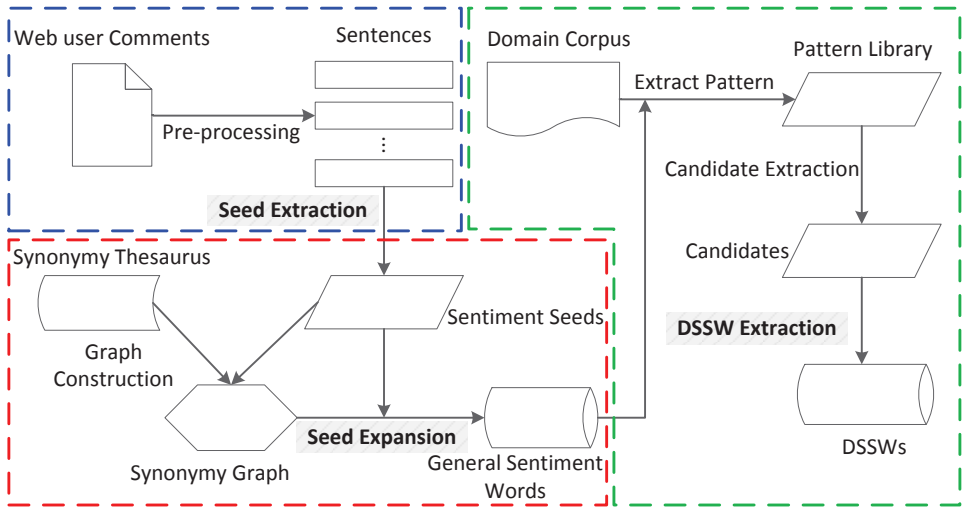}
\vspace{-3em}
\caption{The architecture of proposed framework}
\label{figure-framework}
\end{figure}
\begin{description}
  \item[1.] \textbf{Seed Extraction}: Massive dataset of user comments is used to extract high-confidence sentiment seeds.
  \item[2.] \textbf{Seed Expansion}: First, a synonymy graph is build based on synonyms. Then, graph propagation algorithm is used to get general sentiment words.
  \item[3.] \textbf{DSSW Extraction}: Two kinds of patterns are generated automatically, which are further used to extract new sentiment words from corpus.
\end{description}
\textbf{It's worth noting} that the three-step framework is flexible and we can try different strategies for each component. The following sections give the details algorithms for each component to show its effectiveness and scalability.
\subsection{Seed Extraction}\label{seed extraction}
Sentiment seeds are the foundations of the proposed framework, as shown in Figure \ref{figure-framework}. Most previous work \cite{Turney2003,Qiu2009} manually list a small amount of words as seeds. In this paper, we select sentiment words automatically from ComtData\footnote{ComtData is available at http://www.datatang.com/data/15516}. ComtData includes massive dataset of user comments crawled from a famous Chinese shopping website. It contains 1.65 million user comments on 18K products from 310K users. Each comment includes rating score, overall comments, pros and cons, as illustrated in Table \ref{table-comment}. Pros and cons are used in this paper.

\begin{table}[h]\small
\centering
\begin{tabular}{|c|c|}
\hline
Rating Score & 4.0 \\
\hline
\multirow{2}{*}{Overall comments} & The bag is just what I want! \\
&  However, the logistics disappointed me.\\
\hline
Pros & An excellent bag!\\
\hline
Cons & Disappointing logistics.\\
\hline
\end{tabular}
\caption{Statistics for domain-specific corpora}
\label{table-comment}
\end{table}

Firstly, word segmentation and POS tagging are preprocessed to the pros and cons. All adjectives and idioms are treated as candidates for sentiment seeds. 

Then, two simple schemes are designed to calculate positive and negative score for each seed candidate, namely seed positive score (SPS) and seed negative score (SNS). In Equation \ref{equation_SPS} and \ref{equation_SNS}, $cp_w$ and $cn_w$ denote the frequency of seed candidate \emph{w} in pros and cons. After that, all candidates are ranked by SPS and SNS respectively.
\vspace{-0.5em}
\begin{equation}
SPS(w) = cp_w / (cp_w + cn_w) 
\label{equation_SPS}
\end{equation}
\begin{equation}
SNS(w) = cn_w/ (cp_w + cn_w) 
\label{equation_SNS}
\end{equation}

Finally, sentiment seeds are selected based on the following rules.
\begin{itemize}
\item Word length constraints. Since most of the Chinese single words are ambiguous, we only retain words whose length is greater than 1.
\item Frequency constraints. Seed candidates with frequency smaller than 30 are removed empirically.
\item Sentiment score constraints. After generating the SPS and SNS rankings, the candidates with SPS larger than $\lambda_p$ or SNS larger than $\lambda_n$ are chosen as sentiment seeds, as listed in Table \ref{table_top15seeds} 

\end{itemize}
\begin{CJK*}{UTF8}{gbsn}
\begin{table}[!h]
\centering
	\begin{tabular}{|c|c|c||c|c|c|}
	\hline
	\multicolumn{3}{|c||}{Positive Seeds} & \multicolumn{3}{|c|}{Negative Seeds}\\
	\hline
	Word\_CH/Word\_EN & $c_p$/$c_n$& SPS & Word\_CH/Word\_EN & $c_p$/$c_n$ & SNS\\
	\hline
	古朴/quaint & 32/0 & 1.000 & 臃肿/bloated & 6/47 & 0.887\\
	毋庸置疑/no doubt & 63/1 & 0.984  & 凹凸不平/accidented & 4/30& 0.882 \\
	简捷/simple & 51/1 & 0.981  & 陈旧/obsolete & 26/186& 0.877 \\
	素雅/elegant & 84/2 & 0.977  & 土气/tacky & 6/41 & 0.872 \\
	敏捷/agile & 82/2 & 0.976  & 欠佳/poor & 38/228 & 0.857 \\
	香喷喷/fragrant & 40/1 & 0.976  & 脆弱/fragile & 91/539 & 0.856 \\
	霸气/intrepid & 39/1 & 0.975  & 画蛇添足/superfluous & 5/29 & 0.853 \\
	俱佳/superb & 38/1 & 0.974  & 落后/fall behind & 17/96 & 0.850 \\
	精湛/exquisite & 75/2 & 0.974  & 粗糙/rough & 915/5060 & 0.847 \\
	简练/concise & 37/1 & 0.974  & 费心/exhausting & 6/33 & 0.846 \\
	动听/sweet & 37/1 & 0.974  & 迟钝/obtuse & 52/285 & 0.846 \\
	工细/delicate & 144/4 & 0.973  & 大意/careless & 5/27 & 0.844 \\
	浪漫/romantic & 35/1 & 0.972  & 单薄/flimsy & 531/2797 & 0.840 \\
	优美/beautiful & 266/9 & 0.967 & 累赘/burdensome & 19/100 & 0.840 \\
	一应俱全/complete & 50/2 & 0.962 & 简陋/simple & 446/2322  & 0.839 \\
	\hline
\end{tabular}
\caption{Top 15 sentiment seeds based on SPS and SNS rankings}\label{table_top15seeds}
\end{table}
\end{CJK*}
\subsection{Seed Expansion}\label{graph-based Seed Expansion}
In order to get more domain-independent sentiment words, graph propagation is used to expand sentiment seeds. Firstly, synonymy graph is built with a bootstrapping schema. Then, graph propagation algorithm is utilized on the synonymy graph to expand the sentiment seeds. After the graph propagation converged, top K words are selected as general sentiment words.
\subsubsection{Graph Construction}\label{graph construction}
On the basis of sentiment seeds and Synonymy Thesaurus\footnote{Synonymy Thesaurus is available at http://www.datatang.com/data/13282. Each token in Synonymy Thesaurus has a list of synonyms.}, we use bootstrapping method to construct synonymy graph.
Firstly, all candidates after seed extraction are saved as an origin set. Then, synonyms of the words in the original set will be extracted and added into the set. The bootstrapping process runs iteratively until no more new words can be extracted.

In this way, a synonymy graph $G = <V, E>$ is constructed with $|V|$ nodes and $|E|$ edges. Each node indicates a word, and there exists a edge between two nodes if they are synonymies. The adjacency matrix \textbf{W} indicates the relationship between nodes in G. $W_{ij}$ is calculated by the cosine similarity between the synonyms vectors of $v_i$ and $v_j$, as shown in Equation \ref{consine similarity}. $sv_{ik}$ is a boolean value to indicate whether the k-th word in the vocabulary is the synonym of word $v_i$.

\begin{equation}
W_{ij} = \frac{sv_i \cdot sv_j}{\|sv_i\| \times \|sv_j\|} = \frac{\sum\nolimits_{k=1}^{n}sv_{ik} \times  sv_{jk}}{\sqrt{\sum\nolimits_{k=1}^{n}sv_{ik}^2} \times \sqrt{\sum\nolimits_{k=1}^{n}sv_{jk}^2}}
\label{consine similarity}
\end{equation}

\subsubsection{Graph Propagation}\label{graph propagation}
After graph construction, words in the synonymy graph are connected with their synonymies. In this subsection, we use Multi Topic-Sensitive PageRank algorithm for seed expansion. It's widely accepted that sentiment seeds are good indicators for sentiment expression. What's more, from our observation, words with some specific POS tags are more likely to possess sentiment information, such as adjective and idiom. Thus, we utilize Multi Topic-Sensitive PageRank algorithm on the synonymy graph to calculate the sentiment of each word, in which sentiment seeds and POS tagging information are two relevant topics.

PageRank algorithm \cite{Brin1998} is first proposed to measure the authority of each web page for search result ranking. The idea behind PageRank is that, a page that is linked to by many pages with high rank receives a high rank itself. In this work, the synonymy graph is built based on the sentiment consistency assumption, namely a word that has many positive synonyms receives higher positive score. Thus, PageRank is intuitively reasonable for sentiment seed expansion.

The original PageRank values are iteratively calculated based on Equation \ref{topic_pagerank_vector}, where $\underline{{\bf e}_i = 1/N}$. In Equation \ref{topic_pagerank_vector}, $\underline{\alpha{\bf W}{\bf x}^{k-1}}$ corresponds to the \emph{random walk} operation, and $\underline{(1 - \alpha){\bf e}}$ refers to the \emph{teleport} operation \cite{Manning2008}, $\alpha$ is a damping factor to tradeoff between the two parts, \underline{$x_p^k$} is the pagerank value of webpage p in the k-th iteration. 
In order to derive PageRank values tailored to particular interests, \cite{Haveliwala2003} proposed Topic-Sensitive PageRank, whose main difference from original PageRank is the value of $\textbf{e}$. In original PageRank, each web page has equal probability to be visited in \emph{teleport} operation. However, in Topic-Sensitive PageRank algorithm, the random surfer will teleport to \emph{a random web page on the topic} instead.
\begin{equation}
{\bf x}^k= \alpha{\bf W}{\bf x}^{k-1} + (1 - \alpha){\bf e}
\label{topic_pagerank_vector}
\end{equation}
With regard to the situation that a user has a mixture of interests, for example 60\% sports and 40\% politics, \cite{Manning2008} points that individual's interests can be well-approximated as a linear combination, as shown in Equation \ref{multi topic pr}.
\begin{equation}
{\bf x_t} = \beta {\bf x_{t1}} + (1-\beta){\bf x_{t2}}
\label{multi topic pr}
\end{equation}
In this work, sentiment seeds and certain POS tags are treated as two topics due to their close contact with sentiment. Multi Topic-Sensitive PageRank in Equation \ref{multi topic pr} is used to calculate the sentiment for each word. As for sentiment seeds, we use the parameter $\lambda_p$ and $\lambda_n$ in Section \ref{seed extraction} to control the seed list. As for POS tags, we try different POS tags to observe its influence on Topic-Sensitive PageRank. Finally, $\beta$ is tuned to tradeoff between these two topics.
After graph propagation converges, top K words are selected as general sentiment words. 
\subsection{DSSW Extraction}\label{route-based Domain Lexicon Extraction}
This subsection details the algorithm to extract DSSW based on general sentiment words and domain-specific corpora. Syntactic and sequential patterns are used to represent the relationship between sentiment words and target words. Syntactic pattern is the shortest path from sentiment word to target word in the dependency tree, which indicates the hierarchical information. Sequential pattern is the sequential POS tagging string between sentiment word and target word from left to right, which indicates the plain information.

For example, \emph{excellent} is sentiment word and \emph{camera} is target word in Figure \ref{example-1}. The syntactic pattern of this example is \emph{(S) DE+ ATT+ (T)}, in which \emph{(S)} stands for sentiment word, \emph{(T)} stands for target word, \emph{DE} and \emph{ATT} are the dependency relations along the path on the dependency tree from \emph{(S)} to \emph{(T)}, \emph{+} means the latter word is the father of the former in the dependency structure. The sequential pattern of this example is \emph{(S) a u n (T)}, in which \emph{a u n} is the POS tagging sequence from left to right.
Based on the above observation, DSSW extraction algorithm is summarized in Algorithm \ref{alg:route_domain}.
\renewcommand{\algorithmicrequire}{\textbf{Input:}}
\renewcommand{\algorithmicensure}{\textbf{Output:}}
\begin{algorithm}\caption{DSSW Extraction}\label{alg:route_domain}
\centering
\begin{algorithmic}[1]
\REQUIRE ~~\\
General sentiment  ~~\\ Domain corpus 
\ENSURE ~~\\
Domain-specific sentiment words (DSSWs) ~~\\ ~~\\

\STATE Preprocessing. Word segmentation, POS tagging and dependency parsing are proprocessed on the domain corpus. \label{alg:dc-preprocessing}
\STATE Target words extraction. Nouns will be treated as target words if their frequency is larger than $\gamma_d$. \label{alg:target}
\STATE Pattern extraction. Syntactic patterns and sequential patterns between target words and general sentiment words are extracted automatically. Top $\tau_{syn}$ and $\tau_{seq}$ are selected as pattern library based on frequency. \label{alg:library}
\STATE Candidate sentiment words extraction. \label{alg:candidate} Adjectives and idioms are treated as candidate sentiment words.
\STATE DSSW extraction. Candidate word w will be extracted as DSSW if its syntactic pattern or sequential pattern with target words matches the pattern library. \label{alg:add sent-word}
\end{algorithmic}
\end{algorithm}

After preprocessing (Line \ref{alg:dc-preprocessing}), target words are selected based on word frequency (Line \ref{alg:target}). Then, pattern library is constructed based on the syntactic and sequential relations between target words and general sentiment words (Line \ref{alg:library}). Subsequently, new candidate sentiment words are extracted by matching pattern library. 
Finally, satisfied words will be treated as DSSWs (Line \ref{alg:add sent-word}).
\section{Experiment}\label{experiment}

In this section, three experiments are conducted to verify the effectiveness of our method. Firstly, we evaluate the general sentiment words as a result of seed extraction and seed expansion. Then, based on general sentiment words, DSSWs are extracted in three domain. Finally, the extracted DSSW are applied for sentiment classification application to check its usefulness.

\subsection{Results on General Sentiment Words}
\subsubsection{Dataset and Evaluation Metrics}
General sentiment words are selected by seed extraction and seed expansion, as shown in Figure \ref{figure-framework} . The synonymy graph includes 40,680 nodes and 656K edges. Two annotators are asked to label all these words into positive, negative and neutral. The overall inter annotator agreement is 81.05\%. The distribution of annotated lexicon is shown in Table \ref{table:label lexicon}. We can observe that adjectives and idioms have larger possibility to contain subjective information.

\begin{table}[h]
\centering
	\begin{tabular}{|c|rrrrr|}
	\hline
	\multirow{2}{*}{SENT} & \multicolumn{5}{|c|}{POS tagging information}\\
	\cline{2-6}
 & adj & verb & idiom & noun & other \\
 \hline
	pos & 1,230 & 734 & 1,026 & 266 & 642 \\
	neg & 785 & 904 & 746 & 165 & 797 \\
	neu & 918 & 7,569 & 2,016 & 12,668 & 10,214 \\
	\hline
	sum & 2,933 & 9,207 & 3,788 & 13,099 & 11,653 \\
	\hline
\end{tabular}
\caption{Statistics for Chinese lexicon annotation.}\label{table:label lexicon}
\end{table}

In this paper, P@N metric is used to evaluate the performance of graph propagation \cite{Manning2008}. P@N means the Precision from results within top N rankings.

\subsubsection{Parameter Learning}
In this subsection, we conduct experiments to study the influence of different parameter settings in Multi Topic-Sensitive PageRank. Specifically, damping factor $\alpha$ is used to tradeoff between the \emph{teleport} and \emph{random walk} operation; $\lambda_p$ and $\lambda_n$ are used to control the size of positive and negative seeds; $\beta$ is used to tradeoff between the answers from two topic-specific PageRank. It's worth noting that each parameter is learned by two runs of PageRank, for positive and negative rankings respectively.

\begin{figure}[h]
\centering
\subfigure[Positive PageRank]{\label{figure-pagerank-lambda-pos} \includegraphics[width=0.48\columnwidth]{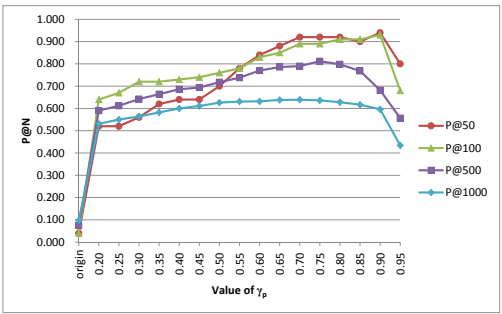} }
\subfigure[Negative PageRank] {\label{figure-pagerank-lambda-neg} \includegraphics[width=0.48\columnwidth]{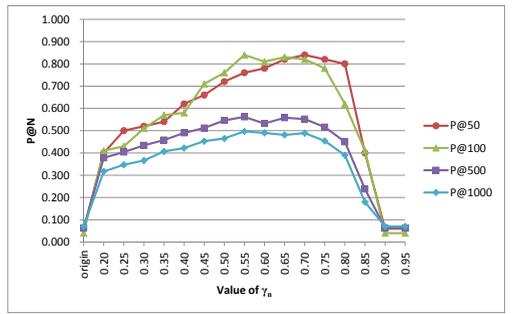} }
\caption{Results on varying values of $\lambda_{pos}$ and $\lambda_{neg}$}
\label{figure-pagerank-lambda}
\end{figure}

Figure \ref{figure-pagerank-lambda} shows the results on varying the value of $\lambda_p$ and $\lambda_n$. The first value \textit{origin} on the horizontal axis means that all nodes are uniformly chosen in \emph{teleport} operation, which corresponds to the origin PageRank. Then, $\lambda_p$ and $\lambda_n$ are increased by 0.05 to control the size of sentiment seeds in Topic-Sensitive PageRank. From Figure \ref{figure-pagerank-lambda-pos} and \ref{figure-pagerank-lambda-neg}, we can observe that Topic-Sensitive PageRank algorithm performs better than the origin PageRank. The best positive and negative PageRank results achieve at $\lambda_p$=0.75 and $\lambda_n$=0.7 respectively. In Figure \ref{figure-pagerank-lambda-neg}, the value at 0.90 and 0.95 are equal to original PageRank value because there are no negative seeds whose SNS value is larger than 0.90.

\begin{figure}[h]
\centering
\subfigure[Positive PageRank]{\label{figure-pagerank-alpha-pos} \includegraphics[width=0.48\columnwidth]{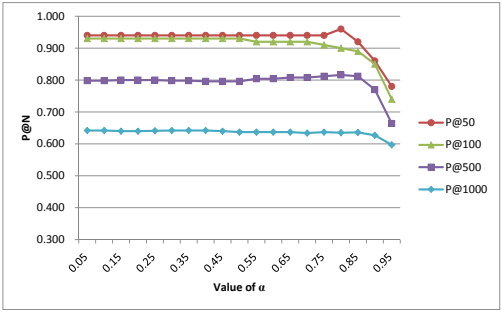} }
\subfigure[Negative PageRank]{\label{figure-pagerank-alpha-neg} \includegraphics[width=0.48\columnwidth]{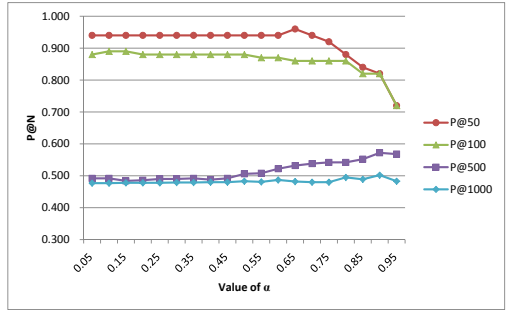} }
\caption{Results on varying values of $\alpha$}
\label{figure-pagerank-alpha}
\end{figure}

Setting $\lambda_p$=0.75 and $\lambda_N$=0.7, the results on varying values of $\alpha$ from 0.05 to 0.95 by 0.05 are shown in Figure \ref{figure-pagerank-alpha}. It's shown that the results change slightly when the value of $\alpha$ is small, where \emph{teleport} operation plays an dominant role in PageRank. However, when $\alpha$ is larger than 0.9, performance drops obviously because the propagation has great probability to conduct \emph{random walk} operation and the effect of sentiment seeds is weaken.

Table \ref{table-postag} shows the effect of adjective, verb and idioms in Topic-Sensitive PageRank. In negative pagerank result, idioms gets the best result. After checking the final ranking result, we find that idioms have more synonymies with other idioms and they have higher probability to act as sentiment word. In addition, the performance in positive PageRank is poor.
\begin{table}[h]\small
\centering
\subtable[Positive PageRank]{
\centering
\begin{tabular}{|c|c|c|c|c|}
\hline
postag & P@50 & P@100 & P@500 & P@1000\\
\hline
i & 0.000  & 0.000  & 0.016  & 0.018 \\
\textbf{a} & \textbf{0.240}  & \textbf{0.280}  & \textbf{0.370}  & \textbf{0.385} \\
v & 0.020  & 0.010  & 0.028  & 0.044\\
\hline
\end{tabular}
\label{table-postag-positive}
}
\subtable[Negative PageRank]{
\centering
\begin{tabular}{|c|c|c|c|c|}
\hline
postag & P@50 & P@100 & P@500 & P@1000 \\
\hline
\textbf{i} & \textbf{0.980} & \textbf{0.960}  & \textbf{0.808}  & \textbf{0.649} \\
a & 0.260  & 0.200  & 0.240  & 0.231 \\
v & 0.020  & 0.040  & 0.032  & 0.048 \\
\hline
\end{tabular}
\label{table-postag-negative}
}
\vspace{-0.5em}
\caption{Results on varying combinations of POS tagging}
\label{table-postag}
\end{table}

\begin{figure}[h]
\centering
\subfigure[Positive PageRank]{\label{figure-pagerank-beta-pos} \includegraphics[width=0.48\columnwidth]{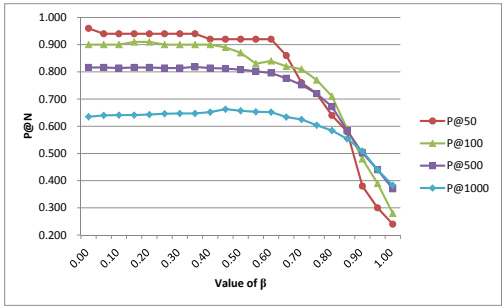} }
\subfigure[Negative PageRank]{\label{figure-pagerank-beta-neg} \includegraphics[width=0.48\columnwidth]{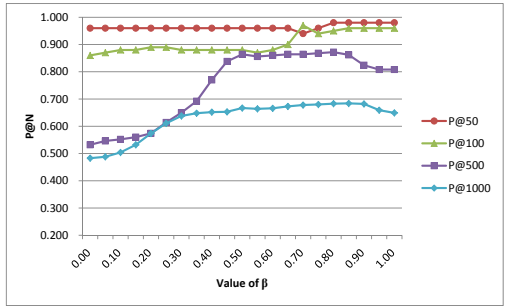} }
\vspace{-0.5em}
\caption{Results on varying values of $\beta$}
\label{figure-pagerank-beta}
\end{figure}

Finally, $\beta$ is set from 0.0 to 1.0 increased by 0.05 to tradeoff between the results of two Topic-Sensitive PageRank methods. Due to the poor performance of positive postag-sensitive PageRank, the best result is given at $\beta$=0.0 in Figure \ref{figure-pagerank-beta-pos}. In Figure \ref{figure-pagerank-beta-neg}, the best result is given around $\beta$=0.75, in which the postag-topic plays a dominant role.

\vspace{-0.5em}
\subsection{Evaluation for DSSW Extraction}\label{experiment-dssl}
\subsubsection{Experiment Setting}
We conduct DSSW extraction on the dataset from Chinese Opinion Analysis Evaluation (COAE 2011) \cite{Zhao2008}. The dataset contains text from three domains, namely digital, entertainment and finance. The detailed information of the corpora is shown in Table \ref{table-statistic-domain-corpus}. Note that the last column means the number of sentiment words (SW) in each domain given by COAE. These sentiment words are considered as gold set in the following experiments. The evaluation metrics are \textbf{P}recision, \textbf{R}ecall and \textbf{F1}-score \cite{Manning2008}.

\begin{table}[h]\small
\centering
\begin{tabular}{|c|c|c|c|}
\hline
domain & \# of docs & \# of sents/doc & \# of SW \\
\hline
finance & \multicolumn{1}{|c|}{14,542} & \multicolumn{1}{|c|}{8} &1,844 \\
entertainment & \multicolumn{1}{|c|}{14,904} & \multicolumn{1}{|c|}{7} & 3,034\\
digital & \multicolumn{1}{|c|}{14,799} & \multicolumn{1}{|c|}{24} & 2,998\\
\hline
\end{tabular}
\caption{Statistics for domain-specific corpora}
\label{table-statistic-domain-corpus}
\end{table}

\subsubsection{Experimental Results}
We re-implement two baselines, \textbf{Hu04} \cite{Hu2004} and \textbf{Qiu11} \cite{Qiu2011} \footnote{The detail of the methods used in baselines will be mentioned in Section \ref{related work}.}. LTP \cite{Che2010} is used for word segmentation, POS tagging and dependency parsing.

In order to compared with the two baselines in the comparable setting, in Algorithm \ref{alg:route_domain}, $\gamma_d$ is set to 100 (Line \ref{alg:target}), $\tau_{syn}$ and $\tau_{seq}$ are both set to 200 (Line \ref{alg:library}). Comparison results on DSSW extraction are given in Table \ref{table:result dssl extraction}.

\begin{table}[h]\small
\centering
\begin{tabular}{|c|c|c|c|c|}
\hline
\multicolumn{2}{|c|}{} & Hu04 & Qiu11 & Our\\
 \hline
\multirow{3}{*}{finance} &P& 0.5423 & 0.5404 & 0.6347\\
\cline{2-5}
 &R& 0.2956 & 0.3118 & 0.3411\\
 \cline{2-5}
 &F1& 0.3826  & 0.3955  & \textbf{0.4437} \\
 \hline
\multirow{3}{*}{entertainment} &P& 0.5626 & 0.5878 & 0.6449\\
\cline{2-5}
 & R&0.2769 & 0.3022 & 0.3256\\
 \cline{2-5}
 & F1&0.3711  & 0.3992  & \textbf{0.4328} \\
 \hline
\multirow{3}{*}{digital} &P& 0.5534 & 0.5649 & 0.5923\\
\cline{2-5}
 & R&0.3043 & 0.3253 & 0.3457\\
 \cline{2-5}
 & F1&0.3927  & 0.4129  & \textbf{0.4366} \\
\hline
\end{tabular}
\caption{Experimental results on DSSW extraction}
\label{table:result dssl extraction}
\end{table}

From Table \ref{table:result dssl extraction}, we observe that our method outperforms two solid baselines in three domains. Our precision(P) improves significantly, especially in finance domain with 9.4\% improvement. Our recall(R) improves slightly because there are still some sentiment words don't co-occur with target words. Problem with hidden target words will be studied in future work.

\begin{figure}[h]
\centering
\subfigure[Results on varying number of patterns]{\label{figure:no-of-routes} \includegraphics[width=0.48\columnwidth]{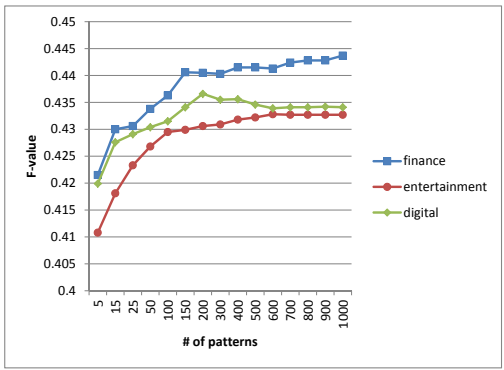} }
\subfigure[Results on varying number of general sentiment words]{\label{figure:no-of-seeds} \includegraphics[width=0.48\columnwidth]{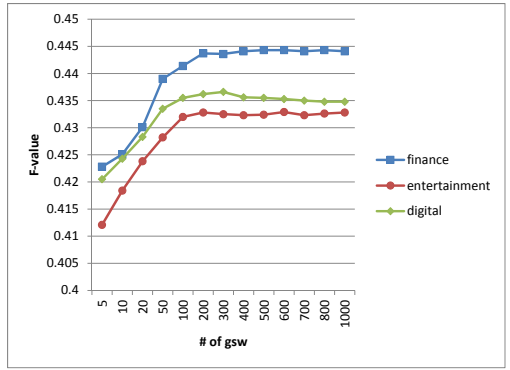} }
\vspace{-0.5em}
\caption{Parameter learning on domain lexicon extraction}
\label{figure-no-of}
\end{figure}

To evaluate the contribution of pattern library and general sentiment words in DSSW extraction, different settings are given. In Figure\ref{figure:no-of-routes}, F-value improves obviously with the increasing size of pattern library within 200. With the expansion of pattern library, new added patterns are not general enough to match mass sentiment words as before. Thus, the trend became stable after 200. In Figure\ref{figure:no-of-seeds}, general sentiment words can be treated as sentiment seeds when its size is tiny. With more general sentiment words added, statistical information of patterns can be learned better. Thus, the performance rises apparently, which indicates the effectiveness of seed expansion. Finally, the trend is stable when the size of general sentiment words is larger than 200.

\subsection{Evaluation for Sentiment Classification}
In order to verify the usefulness of DSSWs extracted by the proposed framework, we apply the DSSWs for sentiment classification task. Motivated by \cite{Pang2002}, machine learning method is used to classify the polarity of text. The objective is to compare the effectiveness of our DSSWs with general sentiment lexicon (HownetSent) and the baseline DSSWs (Hu04, Qiu11) for sentiment classification. We use the dataset from Task2 in COAE2011, which also contains text from three domains. Each sentence in this dataset is labelled with positive or negative. We balance them manually.

\begin{table}[h]\small
\centering
\begin{tabular}{|c|c|c|c|c|}
\hline
 & Hownet & Hu04 & Qiu11 & Our\\
 \hline
finance  & 0.5714  & 0.5543  & 0.5657  & \textbf{0.6029} \\
entertainment & 0.5078  & 0.5596  & 0.5777  & \textbf{0.5962} \\
digital & 0.6800  & 0.6813  & 0.7007  & \textbf{0.7242} \\
\hline
\end{tabular}
\caption{Results on sentiment classification}
\label{tabel:sentiment classification}
\end{table}
As shown in Table \ref{tabel:sentiment classification}, our approach outperforms general sentiment lexicon(HownetSent) and baseline DSSW (Hu04 and Qiu11) in all three domains, which indicates the effectiveness of the extracted DSSWs.

\section{Related Work}\label{related work}
The objective of sentiment word extraction is to identify sentiment words from text. Recently, three main approaches have been investigated: thesaurus-based method, corpus-based method and manual method \cite{Liu2012}. Because the manual method is time-consuming, it's mainly combined with automatic methods as the final check. As follows, we will discuss the thesaurus-based and corpus-based method in detail.
\subsection{Thesaurus-based Method}\label{thesaurus-based method}
Thesaurus-based method mainly utilizes the semantic relation, such as synonyms, antonyms and hypernyms, between tokens in thesaurus (e.g. WordNet) to build general lexicon. Majority of the existing work treat sentiment word as a basic unit\cite{Hu2004}, yet some researchers focus on the synset in WordNet \cite{Esuli2006} and word sense \cite{Wiebe2006,Su2009}.

\cite{Kim2004} propose a simple and effective approach to build lexicon taking advantage of synonym and antonym relations in WordNet. Their hypothesis is that the synonyms of a positive word have positive polarity, and vice versa for antonym relation. In their method, some adjective and verb words are manually labelled as seeds. Then, a bootstrapping method is proposed to expand the seed list. \cite{Kamps2004} utilized synonyms in WordNet to construct a network. The polarity of a word is decided by its shortest path to seed word \emph{good} and \emph{bad}. \cite{Esuli2005} use gloss information to identify the polarity of a word. Their basic assumption is that terms with similar polarity tend to have same glosses. They first collect some seeds manually. Then, a semi-supervised framework is used to classify orientations. Similarly, the work of \cite{Takamura2005} exploit the gloss information to extract polarity of words with spin model. Inspired by \cite{Zhu2002}, \cite{Rao2009} use label propagation algorithm to detect the polarity of words in the graph.

Besides the above-mentioned work, some researchers try to identify the polarity of WordNet synset. \cite{Esuli2006,Baccianella2010} release SentiWordNet, in which each synset is associated with three numerical scores, describing how objective, positive and negative the terms contained in the synset are. Each score in SentiWordNet is in range $\left[0.0, 1.0\right]$ and the summation is 1.0. Similar to \cite{Esuli2005}, their method is also based on quantitative analysis of the glosses associated to synsets. \cite{Esuli2007} utilize pagerank to rank WordNet synsets of how strong they possess positive or negative meaning. Inspired by \cite{Blum2001} and \cite{Pang2004}, \cite{Su2009} propose a semi-supervised mincut framework to recognize the subjectivity of word sense in WordNet. However, the thesaurus-based method can't exploit domain-specific words because most entries in thesaurus (e.g. WordNet) are domain-independent. In addition, the thesaurus-based method doesn't consider the word's behaviour in corpora.

\subsection{Corpus-based Method}\label{corpus-based method}
\cite{Hatzivassiloglou1997} propose the first corpus-based method to extract the polarity of adjective. Their underlying intuition is \emph{sentiment consistency}, namely, words conjoined with AND have the same polarity and words connected by BUT have opposite polarity. Their method starts with a list of sentiment seeds, then some pre-defined conjunction (or conjunction patterns) are used to identify more subjective adjectives together with their polarity. However, this method highly relies on the conjunctions, and it's unable to extract adjectives that are not conjoined. \cite{Turney2003} calculate PMI (point mutual information) and LSA (latent semantic analysis) between candidate words and sentiment seeds to measure their semantic similarity. However, their method is time consuming due to the need for web search result (or huge web-scale corpus). \cite{Hu2004} treat frequency nouns and noun phrases as product feature. In their work, adjectives are extracted as sentiment words if they co-occur with product feature. However, they don't consider the relation between sentiment words and product features.

\cite{Kanayama2006} introduced clause-level \emph{sentiment consistency} to obtain candidates, and a statistical estimation approach is used to pick up appropriate sentiment words. However, the statistical estimation will be unreliable if the corpus is small. Further, \cite{Ding2010} explore intra- and inter- sentence \emph{sentiment consistency} to find domain-specific sentiment words. They show that the same word could even indicate different polarities in the same domain. \cite{Qiu2009,Qiu2011} propose a semi-supervised method named \emph{double propagation} for opinion word expansion and target extraction. They only need an initial opinion lexicon to start the bootstrapping process. The key technique is based on syntactic relations that link opinion words and target words. However, their method requires some pre-defined general syntactic rules between sentiment and target words. \cite{Li2012} combine cross-domain classifier and syntactic relation between sentiment words and target words. But labelled data from source domain is essential to transfer knowledge cross different domains. Our method automatically explore hundreds of syntactic and sequential patterns without any manual work.

\section{Conclusion}\label{conclusion}
This paper presents an automatic framework to construct hundreds of syntactic and sequential patterns for domain-specific sentiment word extraction. Firstly, sentiment seeds are extracted from massive dataset of user comments. Then, general sentiment words are selected by graph propagation. Afterwards, syntactic and sequential patterns are detected automatically with the help of general sentiment words and target words from domain-specific corpora. Finally, new sentiment words will extracted as DSSWs if their structures with target words match the patterns.

Experimental results on three domains show that our method outperforms two solid baselines substantially, especially in precision, which means that our large-scale patterns are precise for sentiment word extraction. With the increasing number of patterns and general sentiment words, the F-value increases obviously. Moreover, the extracted DSSWs outperforms general sentiment lexicon and baseline DSSWs in sentiment classification task, which indicates the usefulness of our method.

In future work, we intend to explore hidden targets to improve the recall of our method. Besides, we plan to rank the extracted patterns to increase the accuracy.

\bibliographystyle{apa}
\bibliography{bibtex}

\end{document}